\documentclass{article} % For LaTeX2e
\usepackage{iclr2027_conference,times}
\iclrfinalcopy

\usepackage{amsmath,amsfonts,bm}

\def\eqref#1{equation~\ref{#1}}
\def\1{\bm{1}}

\DeclareMathAlphabet{\mathsfit}{\encodingdefault}{\sfdefault}{m}{sl}
\SetMathAlphabet{\mathsfit}{bold}{\encodingdefault}{\sfdefault}{bx}{n}

\usepackage{hyperref}
\usepackage{url}

\usepackage{tabularx}
\usepackage{subcaption}
\usepackage{wrapfig}
\usepackage{graphicx}
\usepackage{booktabs}

\title{Test-Time Confidence Adaptation under RL-Induced Trajectory Shift}

\author{%
\makebox[\dimexpr\textwidth-2\tabcolsep\relax][c]{%
\normalfont\small
\begin{tabular}{@{}c@{}}
{\normalsize\bfseries
Zhenzhen Ren\textsuperscript{1,2}\quad
Xiaoqing Zhang\textsuperscript{2,3}\quad
GuoBiao Li\textsuperscript{1}\quad
Bowen Fang\textsuperscript{4}}\\[3pt]
{\normalsize\bfseries
Yanxi Wang\textsuperscript{2,5}\quad
Xinpeng Zhang\textsuperscript{1,6}\quad
Ke Han\textsuperscript{7}}\\[7pt]
\textsuperscript{1}Fudan University\quad
\textsuperscript{2}Zhongguancun Academy\\
\textsuperscript{3}Renmin University of China\\
\textsuperscript{4}Institute of Automation,
Chinese Academy of Sciences\\
\textsuperscript{5}Beijing Normal University\quad
\textsuperscript{6}Shanghai University\\
\textsuperscript{7}Queen Mary University of London
\end{tabular}%
}%
}

\begin{document}
\maketitle
\fancyhead{}
\renewcommand{\headrulewidth}{0pt}

% \maketitle

% At the time of submission, the main text should be 9 pages or fewer. 
% During the discussion/rebuttal phase and for the camera ready, the page limit will be increased to 10 pages to allow for new results/discussions. This limit will be strictly enforced. Papers with main text beyond the page limit will be desk-rejected. 

\begin{abstract}
Trajectory-based confidence estimation has shown promising performance for large language models (LLMs) by leveraging rich signals from their reasoning processes.
However, we find that such confidence estimators can degrade substantially after agentic reinforcement learning (Agentic RL) as it induces substantial shifts in the reasoning trajectories.
More importantly, despite these shifts, we uncover a reliability structure that is largely preserved across Base and RL models: $\textbf{more}$ $\textbf{stable}$ $\textbf{trajectories}$ $\textbf{remain}$ $\textbf{more}$ $\textbf{likely}$ $\textbf{to}$ $\textbf{produce}$ $\textbf{correct}$ $\textbf{answers}$.
This preserved relationship provides a transferable, label-free reliability signal for post-RL confidence adaptation.
Building on this finding, we propose $\textbf{StaR-TTA}$, a stability-guided framework for label-free test-time confidence adaptation.
StaR-TTA jointly learns from correctness supervision and stability-derived reliability supervision on labeled Base trajectories, establishing the connection between trajectory stability and answer reliability.
At test time, it reuses stability-induced reliability rankings from unlabeled RL trajectories to adapt confidence scores to the shifted trajectory distribution, without requiring target correctness labels.
Across four QA benchmarks, StaR-TTA achieves an average AUROC of \(0.721\), with consistent improvements across unseen datasets, model families and scales, and different Agentic RL algorithms.
\end{abstract}

% \begin{abstract}
% Large language models (LLMs) are increasingly deployed after agentic reinforcement learning (Agentic RL) in high-stakes domains such as finance. 
% Reliable confidence estimation is therefore essential. 
% However, existing methods that work well on base LLMs do not transfer directly after the base LLMs are trained by Agentic RL. 
% To address this issue, we first identify the shift in trajectories as the main source of such transfer failure.
% Then, we find the relationship between trajectory stability and correctness are stable even after Agentic RL.
% Based on such finding, we propose StaR-TTA: Label-Free Test-Time Confidence Adaptation via Stability Ranking.
% StaR-TTA first train a Trajectory-aware Confidence Scorer on Base LLM trajectories.
% Then, it adapt its scores on RL trajectories at test time using stability-based rankings, without target labels.
% Across four datasets, StaR-TTA achieves an average AUROC of 0.721 and ECE of 0.069, outperforming  evaluated baselines. 
% Further experiments across multiple LLMs and RL algorithms demonstrate its generalization. 
% These results suggest that StaR-TTA is a practical starting point for reliable confidence estimation in Agentic RL trained LLM agents.
% \end{abstract}

\begin{figure*}[htbp]
    \centering
    \includegraphics[
        width=\textwidth
    ]{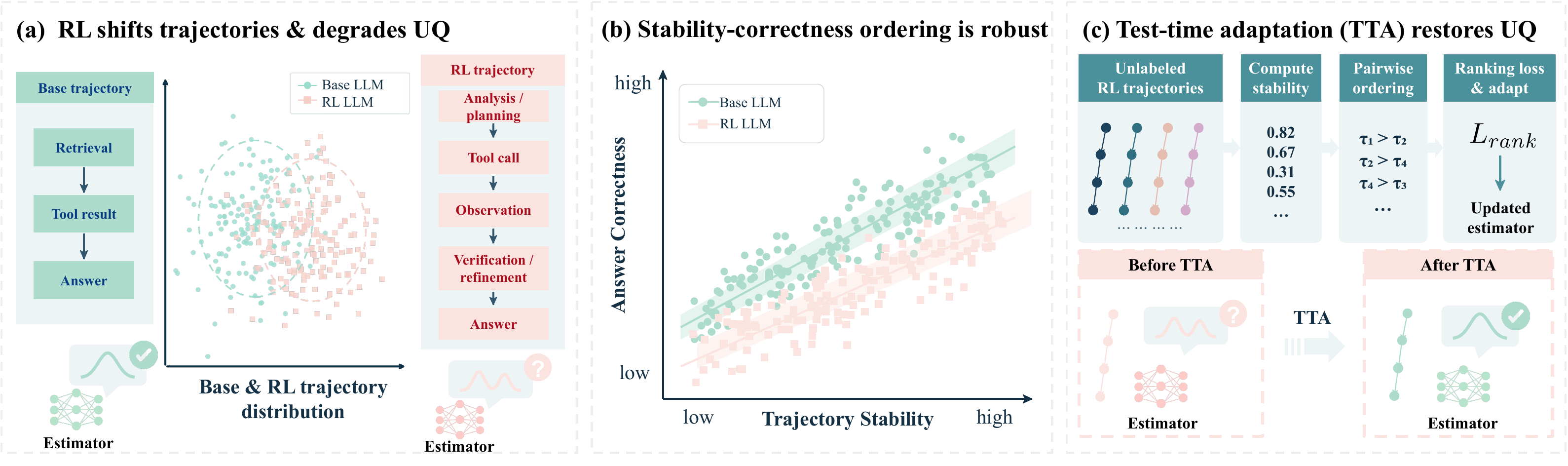}
    \caption{(a) Agentic RL shifts reasoning trajectories and degrades uncertainty quantification (UQ).  (b) Stability-correctness ordering remains highly positive after RL. (c) Stability-guided test-time adaptation restores reliable confidence estimation. Using unlabeled RL trajectories, we exploit trajectory stability to construct pairwise ordering and adapt a confidence estimator at test time with a ranking loss, leading to more reliable confidence estimates.}
    \label{fig:intro}
\end{figure*}

\section{introduction}

Reliable confidence estimation is essential for deploying large language models (LLMs) in real-world applications, particularly for reasoning tasks where plausible outputs can still be incorrect \citep{xiao2025consistency,de2025study,tian2023just,turpin2023language,guan2026beware,xiao2026enhancing}.
Recent approaches increasingly estimate confidence from reasoning trajectories, which reveal how a model arrives at its answer \citep{khanmohammadi2025calibrating,subramani2025mice}.
Meanwhile, LLMs are increasingly optimized through agentic reinforcement learning (Agentic RL), which can substantially reshape their reasoning and tool-use behavior \citep{cui2026rethinking,wei2026agentic,jin2025beneficial,singh2025agentic,matsutani2026rl}.
This raises a critical question:
\textbf{do confidence estimators learned from a base model remain reliable after Agentic RL?}

We find that they often do not.
Confidence estimators that perform well on Base-model trajectories can degrade substantially when directly transferred to their RL-trained counterparts.
Through controlled analysis, we identify post-RL trajectory shift as an important source of this degradation: even with tool results and final answers controlled, replacing RL trajectories with their Base counterparts recovers a substantial portion of the lost confidence-ranking performance.
Retraining requires post-RL correctness labels, while direct transfer suffers from trajectory shift.
We therefore ask: \textbf{can a confidence estimator adapt to post-RL trajectory shift without target correctness labels?}

Our key observation is that, although Agentic RL shifts trajectories, it does not destroy all reliability structure.
Across both Base and RL models, trajectory stability remains positively associated with answer correctness: answers that remain more stable under repeated sampling are consistently more likely to be correct.
Crucially, unlike correctness labels, trajectory stability can be obtained directly from model outputs without human annotation.
This creates a useful supervision bridge across the RL shift:
\textbf{correctness labels can ground the stability-reliability relationship on the source model, while stability remains available as a label-free reliability signal after RL}.

Building on this insight, we propose \textbf{StaR-TTA}, a label-free test-time confidence adaptation framework via \textbf{Sta}bility \textbf{R}anking.
During source training, StaR-TTA jointly learns from correctness supervision and stability-derived reliability supervision, explicitly aligning confidence scores with the reliability ordering induced by trajectory stability.
At test time, correctness labels are unavailable, but the preserved stability--correctness relationship provides a transferable supervisory signal.
StaR-TTA extracts stability from unlabeled RL trajectories, constructs reliability rankings, and uses them to adapt the confidence scorer to the shifted trajectory distribution while preserving the source-learned confidence scale.
The adapted scores are then mapped to correctness probabilities by a frozen confidence calibrator.
In this way, the same stability-based reliability structure learned alongside correctness on the source model becomes the supervision for label-free adaptation after RL.

Extensive experiments on four QA benchmarks show that StaR-TTA consistently improves post-RL confidence estimation, achieving an average AUROC of \(0.721\).
Its gains persist across unseen datasets, different model families and scales, and different Agentic RL algorithms, demonstrating the strong generalization of our method.
% Together, these results show that trajectory shift need not require re-annotating and retraining confidence estimators from scratch: reliability structure preserved across RL can instead provide supervision for test-time adaptation.

Our contributions are summarized as follows:
\begin{itemize}
    \item We reveal that confidence estimation can degrade substantially after Agentic RL, and identify post-RL trajectory shift as an important contributor to the failure of trajectory-based confidence estimators.
    
    \item We uncover a transferable reliability structure under RL-induced trajectory shift: despite substantial changes in reasoning trajectories, the relationship between trajectory stability and answer correctness is largely preserved across Base and RL models.
    
    \item We propose \textbf{StaR-TTA}, a label-free test-time confidence adaptation framework that learns stability-based reliability from source correctness supervision and reuses the preserved reliability ordering to adapt confidence estimation to unlabeled post-RL trajectories.

    \item Extensive experiments show that StaR-TTA consistently improves post-RL confidence estimation and generalizes effectively across models, datasets, and Agentic RL algorithms.
\end{itemize}

\section{Related Work}

\textbf{Confidence Estimation.}
Confidence estimation predicts the probability that an LLM answer is correct \citep{kadavath2022language}.
$P(\mathrm{True})$ provides a self-evaluated signal \citep{kadavath2022language}.
Beyond self-evaluation, CCPS \citep{khanmohammadi2025calibrating} predicts correctness from hidden-state perturbation features, while MICE \citep{subramani2025mice} uses logit-lens comparisons between intermediate and final predictions for tool-using agents.
At the process level, HTC \citep{zhang2026agentic} estimates answer confidence for tool-using agents, whereas $P(\mathrm{Sufficient})$ \citep{zhang2026confidence} probes whether accumulated multi-turn information is sufficient for answering correctly.
DACA \citep{luo2026your} instead calibrates post-trained model confidence through agreement and disagreement with its pre-trained counterpart, without correctness labels.
These methods exploit hidden representations, intermediate predictions, trajectory behavior, and so on, but how reliably these signals transfer under model and trajectory distribution shifts remains an open question.

\textbf{Agentic Reinforcement Learning.}
Agentic reinforcement learning (Agentic RL) trains LLM policies through multi-step interactions with environments, tools, or other agents, using trajectory-level rewards \citep{cui2026rethinking,wei2026agentic}.
Common training algorithms include Proximal Policy Optimization (PPO) and Group Relative Policy Optimization (GRPO) \citep{ouyang2022training,havrilla2024teaching,guo2025deepseek,shao2024deepseekmath}.
Unlike supervised fine-tuning on fixed reasoning traces, Agentic RL lets models learn when to reason, search, verify, and use tools through interaction.
For example, Search-R1 \citep{jin2025search} trains a base LLM to interleave reasoning with search, issue multiple queries, and incorporate retrieved evidence before answering.
It uses multi-turn rollouts, retrieved-token masking, and outcome-based rewards without step-level annotations.
Other work applies Agentic RL to self-reflection and verification \citep{guo2025deepseek}, tool use \citep{singh2025agentic}, and search without supervised reasoning traces \citep{chen2025learning}.
These methods can change action patterns, tool use behaviors, reasoning structure, and hidden representations, shifting the trajectory distributions on which confidence estimators rely.

\section{Confidence Degradation under RL-Induced Trajectory Shift}
\label{sec3}

\subsection{Confidence Estimation Degrades after Agentic RL}

% \noindent
% \textbf{Robustness of Confidence Estimation to RL.}
\begin{wraptable}[11]{r}{0.49\columnwidth}
\vspace{-1em}
\centering
\footnotesize
\setlength{\tabcolsep}{2.5pt}
\renewcommand{\arraystretch}{1.05}
\begin{tabular}{@{}lccc@{}}
\toprule
\textbf{Method}
& \shortstack{Base\\AUROC$\uparrow$}
& \shortstack{RL\\AUROC$\uparrow$}
& \shortstack{Relative\\drop (\%)} \\
\midrule
MICE-LR & 0.741 & 0.567 & 23.5 \\
MICE-RF & 0.773 & 0.559 & 27.7 \\
HTC     & 0.765 & 0.675 & 11.8 \\
CCPS    & 0.745 & 0.491 & 34.1 \\
DACA    & 0.659 & 0.656 & 0.5 \\
$P(\mathrm{Sufficient})$ & 0.627 & 0.689 & $-9.9$ \\
\bottomrule
\end{tabular}
\caption{Base-to-RL confidence transfer on NQ.}
\label{tab:nq_transfer}
\vspace{-0.8em}
\end{wraptable}
We first examine whether confidence estimators learned on a Base LLM remain reliable after Agentic RL.
Table~\ref{tab:nq_transfer} reports AUROC on NQ before and after RL.
MICE-LR, MICE-RF, and CCPS explicitly exploit fine-grained trajectory information, while HTC summarizes structural properties of agent trajectories; DACA and $P(\mathrm{Sufficient})$ rely less directly on trajectory representations.
MICE-LR and MICE-RF drop by 23.5\% and 27.7\%, while CCPS drops by 34.1\%.
In contrast, DACA remains nearly unchanged and $P(\mathrm{Sufficient})$ improves.
These results reveal a clear vulnerability: confidence estimators that exploit trajectory information can suffer substantial degradation in confidence ranking after transfer to an RL-trained model.
This motivates us to examine whether changes in the underlying trajectories are responsible for the degradation. 
% HTC uses structural features of trajectories rather than fine-grained representations.
% Under Base LLM, the trajectory-based methods achieve stronger AUROC performance, showing that the trajectory signal is an effective cue for confidence estimation.
% After transfer to the Agentic RL LLM, however, MICE, CCPS exhibit substantial degradation. 
% In contrast, DACA remains robust, while $P(\mathrm{Sufficient})$ even improves. 
% These results suggest that trajectory-based confidence estimators, despite competitive performance on the base model, are particularly sensitive to Agentic RL training.Trajectory-based estimators perform strongly on the Base model but degrade markedly after RL.

\subsection{Isolating the Effect}

% \begin{wraptable}[7]{r}{0.49\columnwidth}
% \vspace{-1.5em}
% \centering
% \footnotesize
% \setlength{\tabcolsep}{2.0pt}
% \renewcommand{\arraystretch}{1.08}
% \begin{tabular}{@{}lrrrr@{}}
% \toprule
% \textbf{Method}
% & \shortstack{Tool\\control}
% & \shortstack{Answer\\control}
% & \shortstack{Trajectory\\replay}
% & \shortstack{Recovery\\(\%)} \\
% \midrule
% MICE-LR & $-0.174$ & $-0.110$ & $+0.049$ & 44.5 \\
% MICE-RF & $-0.215$ & $-0.087$ & $+0.046$ & 52.9 \\
% CCPS    & $-0.255$ & $-0.133$ & $+0.015$ & 11.3 \\
% \bottomrule
% \end{tabular}
% \caption{Controlled AUROC analysis on NQ.}
% \label{tab:nq_controls}
% \vspace{-0.8em}
% \end{wraptable}

\begin{wraptable}[5]{r}{0.55\columnwidth}
\vspace{-4.5em}
\centering
\footnotesize
\setlength{\tabcolsep}{2pt}
\renewcommand{\arraystretch}{1.2}
\begin{tabular}{@{}lcccc@{}}
\toprule
Method
& \shortstack{Tool\\control}
& \shortstack{Answer\\control}
& \shortstack{Trajectory\\replay}
& \shortstack{Recovery\\(\%)} \\
\midrule
MICE-LR
& $.567\,{\scriptscriptstyle(-.174)}$
& $.514\,{\scriptscriptstyle(-.110)}$
& $.563\,{\scriptscriptstyle(+.049)}$
& 44.5 \\
MICE-RF
& $.559\,{\scriptscriptstyle(-.215)}$
& $.577\,{\scriptscriptstyle(-.087)}$
& $.623\,{\scriptscriptstyle(+.045)}$
& 51.9 \\
CCPS
& $.491\,{\scriptscriptstyle(-.255)}$
& $.498\,{\scriptscriptstyle(-.133)}$
& $.513\,{\scriptscriptstyle(+.015)}$
& 11.1 \\
\bottomrule
\end{tabular}
\caption{Controlled AUROC analysis on NQ.}
\label{tab:nq_controls}
\vspace{-0.8em}
\end{wraptable}

To isolate the source of this degradation, we progressively control three factors that can change after Agentic RL: tool results, final answers, and reasoning trajectories.
The results are summarized in Table~\ref{tab:nq_controls}.

Under \textit{tool control}, we replace RL tool results with their corresponding Base-model results, yet substantial AUROC gaps remain.
Under \textit{answer control}, we further restrict evaluation to examples for which the Base and RL models produce identical tool results and final answers. The degradation still persists.
Finally, under \textit{trajectory replay}, we keep the controlled tool results and final answers fixed but replace the RL reasoning trajectory with its Base-model counterpart.
This recovers 44.5\%, 51.9\%, and 11.1\% of the remaining AUROC loss for MICE-LR, MICE-RF, and CCPS, respectively.
These results show that changes in tools or final answers alone cannot explain the transfer failure.
Instead, the shift in reasoning trajectories itself contributes substantially to the degradation of trajectory-based confidence estimation.

This raises a key challenge: if the trajectory distribution changes after RL, what reliability information remains transferable from the Base model?

% To identify the source of the post-RL degradation, we progressively control the
% tool call results, final answer, and reasoning trajectory, with the results
% shown in Table~\ref{tab:nq_controls}. 
% Under \textit{tool control}, we replace the tool results with the corresponding Base tool results, but a substantial AUROC gap remains. 
% Under \textit{answer control}, we further restrict the analysis to examples with identical tool results and final answers from the Base and RL models; the performance gap persists. 
% Finally, under \textit{trajectory replay}, we keep the tool results and final answers fixed while replacing the RL reasoning trajectory with the corresponding Base trajectory. This replay recovers 44.5\%, 52.9\%, and 11.3\% of the AUROC loss for MICE-LR, MICE-RF, and CCPS, respectively. 
% These controlled experiments indicate that reasoning-trajectory shift accounts for a substantial fraction of the confidence degradation after RL.

\subsection{Stability–Correctness Association Is Largely Preserved after Agentic RL}

Despite the substantial trajectory shift, we find that not all reliability information is lost.
In particular, the relationship between \textbf{trajectory stability} and answer correctness remains informative after Agentic RL.
For each trajectory, we fix the prefix before final-answer generation and only sample multiple short answers without repeating interactions.
We characterize their stability using three complementary signals:
$\mathcal S_{\mathrm{GA}}$, the fraction of sampled short answers matching the original answer;
$\mathcal S_{\mathrm{EC}}$, the concentration of the sampled short answer distribution; and $\mathcal S_{\mathrm{FG}}$, the frequency gap between the two most common answers.
We further consider their equal-weight average, $\mathcal S_{\mathrm{raw3}}$.

\begin{figure*}[t]
    \centering
    \includegraphics[width=0.97\textwidth]{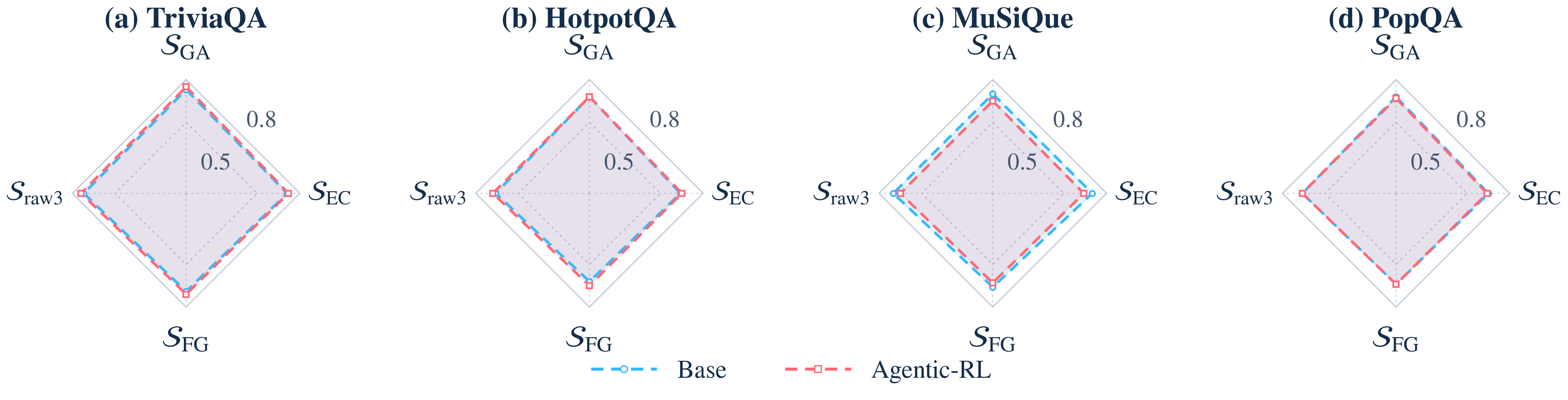}

    \caption{AUROC of sampling-stability signals for answer correctness.}
    \label{fig:stability_correctness}
\end{figure*}

Figure~\ref{fig:stability_correctness} evaluates each stability signal as a predictor of answer correctness on both Base and RL trajectories.
Across all four datasets, the stability signals achieve AUROC well above 0.5 under both Base and Agentic-RL models, while $\mathcal S_{\mathrm{raw3}}$ consistently reaches around 0.7–0.8. This predictive ability is largely retained after Agentic RL despite substantial trajectory shifts.
Thus, despite substantial trajectory shifts after Agentic RL, more stable trajectories remain more likely to produce correct answers.
This observation provides a transferable reliability signal across the RL-induced trajectory shift.
Unlike correctness labels, trajectory stability can be obtained directly from unlabeled model outputs.
This motivates us to leverage stability-derived reliability ordering as label-free supervision to adapt confidence estimation to post-RL trajectories.

\begin{figure*}[t]
    \centering
    
    \includegraphics[
    width=0.98\textwidth,
    trim={0pt 0pt 0pt 0pt},
    clip
]{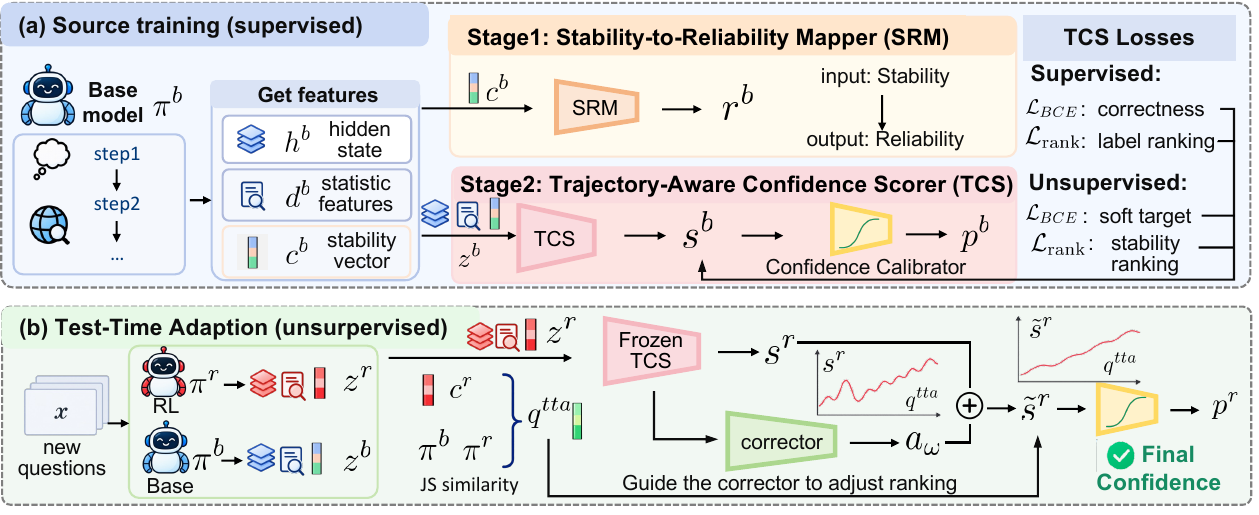}
    \caption{Overview of our method. (a) Firstly, an SRM model is trained to capture the relationship between trajectory stability and correctness. Then a scorer (TCS) is optimized, followed by a calibrator (CC) to convert scores into calibrated probabilities.
    (b) At test time, we adapt the results of TCS to Agentic-RL trajectories. We extract stability signals from both Agentic-RL and Base trajectories and use this information to adjust the score while keeping TCS and CC fixed .}
    \label{fig:method}
\end{figure*}

% We first extract stability signals from both trajectories and use them to adjust the frozen scorer prediction, then apply the frozen calibrator to obtain calibrated confidence scores.
% We represent each trajectory $\tau_i^m$ using internal, external, and stability signals. 
% The internal representation $h_i^m\in\mathbb{R}^{64}$ is extracted from the answer-generation position.
% It summarizes the information accumulated through reasoning and tool interactions. 
% The external descriptor $d_i^m\in\mathbb{R}^{24}$ captures event and tool-call counts, statistics of tool inputs and returns, invalid actions, and so on.
% These features describe how the answer was produced. 
% For stability, we fix the trajectory prefix before final answer and sample five short answers. 
% This adds little computational overhead because sampling them only costs dozens of tokens.
% After normalizing the answer strings, we compute

\section{Method}

Our analysis in Sec.~\ref{sec3} shows that Agentic RL substantially shifts trajectory representations, while trajectory stability remains predictive of correctness. 
Based on this finding, we propose \textbf{StaR-TTA}, which uses stability as a transferable reliability signal for post-RL confidence adaptation.

As illustrated in Figure~\ref{fig:method}, StaR-TTA jointly learns from correctness and stability-derived reliability supervision on labeled Base trajectories. At test time, stability-induced rankings provide label-free supervision for adapting confidence scores to RL trajectories, mitigating the impact of trajectory shift on confidence estimation.
%which are subsequently converted into probabilities by a frozen calibrator.

% As illustrated in Figure~\ref{fig:method}, our method contains two part. 
% In the first part, we train a Stability-to-Reliability Mapper(SRM) to capture the relationship between trajectory stability and correctness, then we optimize a Trajectory-aware Confidence Scorer(TCS), and finally train a confidence calibrator(CC) to convert the TCS output into probabilities. 
% In the second part, we adapt the TCS to Agentic-RL trajectories with unlabeled paired Base and RL trajectories at test time. 

\subsection{Problem Setup}

% We study confidence transfer from a Base LLM $M_b$ to its Agentic-RL counterpart $M_r$. Given a question $q_i$, model $M_m$, where $m\in\{b,r\}$, generates an interaction trajectory $\tau_i^m=M_m(q_i)$ containing its reasoning process, tool interactions, and final answer $a_i^m$. Let $y_i^m\in\{0,1\}$ denote the correctness of $a_i^m$, and let $p_i^m=P(y_i^m=1\mid q_i,\tau_i^m)$ denote its confidence. During source training, correctness labels $y_i^b$ are available for Base-model trajectories, whereas target correctness labels $y_i^r$ are unavailable at test time. Our goal is to adapt a Base-trained confidence estimator to the post-RL trajectory distribution using only signals derived from unlabeled target trajectories.

We study confidence transfer from a Base LLM $M_b$ to its Agentic-RL counterpart $M_r$. 
Given a question $q_i$ (hereafter, we omit the subscript $i$ unless otherwise specified), model $M_m$, where $m\in\{b,r\}$, generates an interaction trajectory $\tau^m=M_m(q)$ containing its reasoning process, tool interactions, and final answer $a^m$. 
Let $y^m\in\{0,1\}$ denote the correctness of $a^m$, and let $p^m=P(y^m=1\mid q,\tau^m)$ denote its confidence. 
During source training, correctness labels $y^b$ are available for Base-model trajectories, whereas target correctness labels $y^r$ are unavailable at test time.
Our goal is to adapt a Base-trained confidence estimator to the post-RL trajectory distribution using only signals derived from unlabeled target trajectories.

% We study the transfer of confidence estimation from Base LLMs to their Agentic-RL counterparts. 
% Let $M_b$ and $M_r$ denote a frozen Base LLM and its Agentic-RL version.
% Given a question $q_i\in\mathcal{D}$, the two models generate corresponding trajectories $\tau_i^m=M_m(q_i)$, where $m\in\{b,r\}$. 
% Each trajectory contains the agent's reasoning process, tool interactions, and the final answer.
% We denote the internal hidden states during the interaction as $H_i^m=\{h_{i,1}^m,\ldots,h_{i,T}^m\}$ for each trajectory $\tau_i^m$, and the final answer as $a_i^m$.
% A confidence estimator aims to predict whether the generated answer is
% correct based on the question and its trajectory, i.e.,
% $p_i^m$, which estimates the probability
% $P(y_i^m=1|q_i,\tau_i^m)$ and $y_i^m$ indicates the correctness of the final
% answer.
% Our goal is to transfer confidence estimation learned from Base-LLM trajectories to Agentic-RL trajectories, without relying on correctness labels.

\subsection{Stability-Grounded Source Confidence Learning}
\label{sec:source_training}

% \paragraph{Trajectory representation.}
% We represent each trajectory using internal, interaction, and stability features. We extract a $64$-D hidden representation $h_i^m$ at the answer-generation position and a $24$-D descriptor $d_i^m$ of trajectory statistics, including tool interactions, invalid actions, and reasoning-related statistics. To measure stability without repeated agent rollouts, we fix the trajectory prefix before final-answer generation and sample five short answers. We compute
% \[
% c_i^m=[\mathrm{Supp}_i^m,\;1-\mathcal H_i^m,\;\mathrm{FSD}_i^m]\in\mathbb R^3,
% \]
% where $\mathrm{Supp}$ measures agreement with the greedy answer, $\mathcal H$ is normalized answer entropy, and $\mathrm{FSD}$ is the frequency gap between the two most common answers. The complete representation is
% \[
% z_i^m=[h_i^m;d_i^m;c_i^m]\in\mathbb R^{91}.
% \]

\paragraph{Trajectory representation.}

We represent each trajectory using internal, interaction, and stability features.
We extract a $64$-D hidden representation $h^m$ at the answer-generation position and a $24$-D descriptor $d^m$ of trajectory statistics, including tool interactions, invalid actions, and other related statistics.
To measure stability without repeated agent rollouts, we fix the trajectory prefix before final-answer generation and sample five short answers. After answer normalization, we compute support $\mathcal S_{\mathrm{GA}}^m=f^m(a^m)$, entropy concentration $\mathcal S_{\mathrm{EC}}^m=1+\sum_{a:f^m(a)>0}f^m(a)\log f^m(a)/\log 5$, and frequency gap $\mathcal S_{\mathrm{FG}}^m=f_{(1)}^m-f_{(2)}^m$, where $f^m(a)$ is the empirical answer frequency, $a^m$ is the original final answer, and $f_{(1)}^m,f_{(2)}^m$ are the two largest frequencies (zero-padded).
Let $c^m=[\mathcal S_{\mathrm{GA}}^m,\mathcal S_{\mathrm{EC}}^m,\mathcal S_{\mathrm{FG}}^m]\in\mathbb R^3$.
The complete representation is
\[
z^m=[h^m;d^m;c^m]\in\mathbb R^{91}.
\]

\paragraph{Stability-to-Reliability Mapper.}
Stability is available without correctness labels but provides only an indirect reliability signal. We therefore use labeled Base trajectories to learn a Stability-to-Reliability Mapper (SRM),
\[
r^b=M_\phi(c^b)=\sigma(w^\top c^b+b),
\qquad
\mathcal L_{\mathrm{SRM}}=\mathrm{BCE}(r^b,y^b).
\]
The SRM learns how the three stability signals relate to correctness.

\paragraph{Trajectory-Aware Confidence Scorer.}
% Here, we reintroduce the sample indices $i$ and $j$ to distinguish trajectories in pairwise ranking.
To estimate the confidence from trajectory signals, we train a Trajectory-Aware Confidence Scorer (TCS).
\[
s^b=g_\theta(z^b)
\]
using both correctness and SRM-derived reliability supervision. Correctness provides direct supervision through BCE and pairwise ranking, while the SRM provides a transferable reliability ordering.
We use index $i$ and $j$ to represent different source trajectories:
\[
\mathcal P_y=\{(i,j):y_i^b=1,\;y_j^b=0\},
\qquad
\mathcal P_r=\{(i,j):r_i^b-r_j^b>\delta\}.
\]
For either pair set $\mathcal P$, we define
\[
\mathcal L_{\mathrm{rank}}(\mathcal P)
=
\frac{1}{|\mathcal P|}
\sum_{(i,j)\in\mathcal P}
\log(1+\exp[-(s_i^b-s_j^b)]).
\]
The source objective is
\[
\mathcal L_{\mathrm{score}}
=
\underbrace{
\mathrm{BCE}(\sigma(s_i^b),y_i^b)
+\lambda_h\mathcal L_{\mathrm{rank}}(\mathcal P_y)
}_{\text{correctness supervision}}
+
\underbrace{
\lambda_r\mathcal L_{\mathrm{rank}}(\mathcal P_r)
+\lambda_s\mathrm{BCE}(\sigma(s_i^b),r_i^b)
}_{\text{stability-derived supervision}}.
\]
This joint training connects correctness-based confidence estimation with the stability-derived reliability signal available after RL.

Finally, we freeze the TCS and fit a monotonic Confidence Calibrator (CC) on Base-model scores and correctness labels:
\[
p_i^b=C_\psi(s_i^b)=\sigma(\alpha s_i^b+\beta), \qquad \alpha>0,
\]
This converts the TCS score into a probability of correctness while preserving its ranking. 

\subsection{Stability-Guided Test-Time Confidence Adaptation}
\label{sec:tta}

Directly applying the Base-trained TCS to RL trajectories yields $s^r=g_\theta(z^r)$, whose reliability can degrade under RL-induced trajectory shift. Although trajectory stability remains predictive of correctness, it is only an indirect reliability signal and cannot serve as a calibrated confidence estimate. We therefore use stability as \emph{adaptation supervision}: it specifies how confidence scores should be reordered after RL, while the source-trained TCS provides the initial confidence estimate.

\paragraph{Target reliability ranking.}
RL stability alone characterizes the consistency of the post-RL model, but does not capture how strongly its behavior has changed from the Base model. We therefore complement $c^r$ with Base--RL answer agreement. Given sampled-answer distributions $\pi^b$ and $\pi^r$, we compute
\[
A=1-\frac{\mathrm{JSD}(\pi^b,\pi^r)}{\log 2},
\]
and construct the label-free target reliability signal
\[
q^{\mathrm{tta}}
=
\frac{\mathcal S_{\mathrm{GA}}^r+\mathcal S_{\mathrm{EC}}^r+\mathcal S_{\mathrm{FG}}^r+A}{4}.
\]
Rather than treating $q^{\mathrm{tta}}$ as confidence, we use it only to infer relative reliability.
Specifically,
\[
\mathcal P_{\mathrm{tta}}
=
\{(i,j):q_i^{\mathrm{tta}}-q_j^{\mathrm{tta}}>\delta_{\mathrm{tta}}\},
\qquad
t_{ij}
=
\sigma\!\left(
\frac{q_i^{\mathrm{tta}}-q_j^{\mathrm{tta}}}{T}
\right),
\]
where $i,j$ index different target trajectories and $t_{ij}$ emphasizes larger reliability gaps.

\paragraph{Trajectory-conditioned residual adaptation.}
A direct correction based only on $q^{\mathrm{tta}}$ would apply similar adjustments to samples with similar stability, despite potentially different trajectory shifts. Instead, we use the stability-induced ordering to supervise an \emph{instance-dependent} correction conditioned on the TCS trajectory representation. Specifically, we freeze the TCS and optimize a lightweight linear corrector $a_\omega$:
\[
\tilde s^r=s^r+a_\omega(e^r),
\]
where $e^r$ is the intermediate TCS representation. The residual formulation preserves the source-trained score while allowing trajectory-specific adaptation.

We use $i,j$ to index trajectories and optimize
\[
\mathcal L_{\mathrm{tta}}
=
\underbrace{
\frac{1}{|\mathcal P_{\mathrm{tta}}|}
\sum_{(i,j)\in\mathcal P_{\mathrm{tta}}}
\mathrm{BCE}\!\left(
\sigma(\tilde s_i^r-\tilde s_j^r),t_{ij}
\right)
}_{\text{stability-guided adaptation}}
+
\underbrace{
\frac{\lambda_a}{N}
\sum_i(\tilde s_i^r-s_i^r)^2
}_{\text{source-score preservation}}.
\]
The ranking term adapts confidence ordering to the target reliability structure, whereas the anchoring term prevents noisy stability supervision from overriding the correctness information learned on the source model.
Only $a_\omega$ is updated during adaptation, while both LLMs, the SRM, TCS, and CC remain frozen. After adapting $a_\omega$ on a small set of unlabeled target samples, we freeze the corrector and apply the resulting confidence estimator to all subsequent RL trajectories without further optimization. For a new trajectory, the adapted score and final confidence are computed as
\[
\tilde s^r=s^r+a_{\omega^\ast}(e^r),
\qquad
p^r=C_\psi(\tilde s^r),
\]
where $\omega^\ast$ denotes the adapted corrector parameters.

\section{Experiments}

\subsection{Experimental Setup}
We use Qwen2.5-3B \citep{qwen2024qwen25} as our test model and Search-R1 \citep{2025Search} as the Agentic RL framework. 
% We diagnose Base-to-RL confidence degradation on NQ \citep{2019Natural} and 
We evaluate effectiveness on TriviaQA \citep{2017TriviaQA}, HotpotQA \citep{2018HotpotQA}, MuSiQue \citep{2021MuSiQue}, and PopQA \citep{mallen2023when}. For cross-dataset generalization, we train on TriviaQA and test on NQ, 2WikiMultiHopQA \citep{2020Constructing}, and Bamboogle \citep{2023Measuring}. 
We also test different model scales and families using Qwen2.5-7B-Instruct \citep{qwen2024qwen25} and Llama-3.2-3B-Instruct \citep{meta2024llama32};
and different RL algorithms: PPO \citep{havrilla2024teaching}, GRPO \citep{guo2025deepseek}.
% We further test model and algorithm robustness using Qwen2.5-7B-Instruct \citep{qwen2024qwen25} and Llama-3.2-3B-Instruct \citep{meta2024llama32} with PPO algorithm \citep{havrilla2024teaching}, and Qwen2.5-3B with GRPO \citep{guo2025deepseek} algorithm. 
% Unless otherwise stated, target correctness labels are used only for evaluation.
We compare with five baselines: CCPS \citep{khanmohammadi2025calibrating}, MICE \citep{subramani2025mice}, HTC \citep{zhang2026agentic}, P(Sufficient) \citep{zhang2026confidence}, and DACA \citep{luo2026your}. 
MICE-LR and MICE-RF share internal features but use logistic regression and random forests, respectively.
We report AUROC for ranking, ECE for calibration, and Brier score and NLL for probability quality \citep{2005An,2017On,1950VERIFICATION,2007Strictly}.
We set \(\lambda_a = 0.1\), $\lambda_{\mathrm{h}}=0.5$, $\lambda_r=1.0$, and $\lambda_s=0.5$ (details can be accessed in Appendix~\ref{app:anchor_weight} and Appendix~\ref{app:train_hy}).

\begin{table}[t]
\centering
\caption{Main results of confidence estimation on the RL model's test trajectories.}
\label{tab:main}
\fontsize{10}{11}\selectfont
\setlength{\tabcolsep}{1.25pt}
\renewcommand{\arraystretch}{1.0}
\resizebox{\linewidth}{!}{%
\begin{tabular}{@{}lcccc@{\hspace{5pt}}cccc@{\hspace{5pt}}cccc@{\hspace{5pt}}cccc@{}}
\toprule
& \multicolumn{4}{c}{TriviaQA}
& \multicolumn{4}{c}{HotpotQA}
& \multicolumn{4}{c}{MuSiQue}
& \multicolumn{4}{c}{PopQA} \\
\cmidrule(lr){2-5}
\cmidrule(lr){6-9}
\cmidrule(lr){10-13}
\cmidrule(lr){14-17}
Method
& {\footnotesize AUROC\textsuperscript{$\uparrow$}}
& {\footnotesize ECE\textsuperscript{$\downarrow$}}
& {\footnotesize Brier\textsuperscript{$\downarrow$}}
& {\footnotesize NLL\textsuperscript{$\downarrow$}}
& {\footnotesize AUROC\textsuperscript{$\uparrow$}}
& {\footnotesize ECE\textsuperscript{$\downarrow$}}
& {\footnotesize Brier\textsuperscript{$\downarrow$}}
& {\footnotesize NLL\textsuperscript{$\downarrow$}}
& {\footnotesize AUROC\textsuperscript{$\uparrow$}}
& {\footnotesize ECE\textsuperscript{$\downarrow$}}
& {\footnotesize Brier\textsuperscript{$\downarrow$}}
& {\footnotesize NLL\textsuperscript{$\downarrow$}}
& {\footnotesize AUROC\textsuperscript{$\uparrow$}}
& {\footnotesize ECE\textsuperscript{$\downarrow$}}
& {\footnotesize Brier\textsuperscript{$\downarrow$}}
& {\footnotesize NLL\textsuperscript{$\downarrow$}} \\
\midrule
DACA
& .668 & \textbf{.027} & .223 & .638
& .613 & .155 & .243 & .679
& .625 & .310 & .229 & .650
& .680 & .110 & .244 & .681 \\
P(Sufficient)
& .661 & .331 & .337 & 1.131
& .636 & .526 & .498 & 1.551
& .648 & .643 & .551 & 1.584
& .641 & .459 & .446 & 1.382 \\
HTC
& .653 & .289 & .316 & .989
& .507 & .175 & .268 & .738
& \textbf{.721} & .081 & \textbf{.132} & \textbf{.423}
& .678 & .328 & .334 & .937 \\
MICE-LR
& .675 & .125 & .232 & .670
& .640 & .112 & .234 & .662
& .623 & \textbf{.054} & .135 & .443
& .595 & .282 & .321 & .882 \\
MICE-RF
& .629 & .046 & .229 & .651
& .615 & .048 & .219 & .629
& .610 & .064 & .136 & .442
& .590 & .119 & .254 & .702 \\
CCPS
& .420 & .405 & .429 & 1.335
& .438 & .205 & .284 & .920
& .494 & .114 & .149 & .534
& .425 & .306 & .357 & 1.407 \\
\midrule
Ours
& \textbf{.800} & .054 & \textbf{.181} & \textbf{.542}
& \textbf{.715} & \textbf{.041} & \textbf{.200} & \textbf{.582}
& .672 & .120 & .145 & .532
& \textbf{.696} & \textbf{.061} & \textbf{.223} & \textbf{.635} \\
\bottomrule
\end{tabular}%
}
\end{table}

\begin{figure*}[t]
    \centering
    \vspace{0em}
    \includegraphics[
        width=0.95\textwidth
    ]{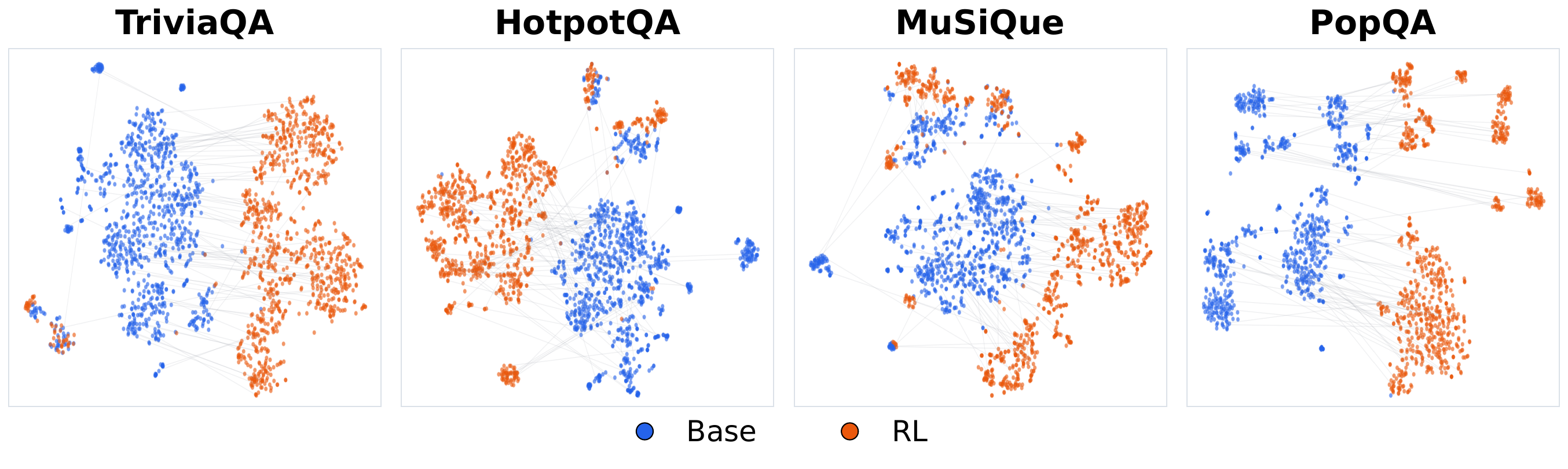}
    \vspace{-1em}
    \caption{
        t-SNE visualization of the H64 hidden representations for Base and RL. 
    }
    \label{fig:hidden-tsne}
\end{figure*}

\subsection{Main Results}

We evaluate confidence transfer in Table~\ref{tab:main}. 
For each dataset, we train the confidence estimators on Base-LLM trajectories and evaluate them on Agentic RL trajectories. 
Across four datasets, our method achieves the best average performance: AUROC ($0.721$), ECE ($0.069$), Brier ($0.187$), and NLL ($0.573$).
In terms of AUROC, only HTC outperforms our method on the MuSiQue dataset.
Appendix~\ref{app:base_rl_test_comparison} compares the performance on both Base and RL trajectories, indicating that the poor performance of baselines is mainly due to the shift of Agentic RL.
We further visualize the representation shift induced by Agentic RL. 
Fig.~\ref{fig:hidden-tsne} shows that the Base and RL representations occupy visibly different regions of the feature space, indicating that RL training alters the internal representation adopted by the confidence estimator.
We further report the baseline that directly uses stability information; the results are shown in Appendix~\ref{app: simple_baseline}, which indicate that stability provides useful information but stability alone is insufficient.
These results show that our method improves both confidence ranking and probability estimation under the influence of Agentic RL.

\subsection{Generalization Analysis}

\subsubsection{Generalization across Datasets}

\begin{table*}[t]
\centering
\caption{Cross-dataset generalization of confidence estimation.}
\label{tab:cross_dataset}
\fontsize{10}{11}\selectfont
\setlength{\tabcolsep}{1.25pt}
\renewcommand{\arraystretch}{1.0}
\resizebox{\linewidth}{!}{%
\begin{tabular}{@{}lcccc@{\hspace{5pt}}cccc@{\hspace{5pt}}cccc@{\hspace{5pt}}cccc@{}}
\toprule
& \multicolumn{4}{c}{TriviaQA}
& \multicolumn{4}{c}{2WikiMultiHopQA}
& \multicolumn{4}{c}{Bamboogle}
& \multicolumn{4}{c}{NQ} \\
\cmidrule(lr){2-5}
\cmidrule(lr){6-9}
\cmidrule(lr){10-13}
\cmidrule(lr){14-17}
Method
& {\footnotesize AUROC\textsuperscript{$\uparrow$}}
& {\footnotesize ECE\textsuperscript{$\downarrow$}}
& {\footnotesize Brier\textsuperscript{$\downarrow$}}
& {\footnotesize NLL\textsuperscript{$\downarrow$}}
& {\footnotesize AUROC\textsuperscript{$\uparrow$}}
& {\footnotesize ECE\textsuperscript{$\downarrow$}}
& {\footnotesize Brier\textsuperscript{$\downarrow$}}
& {\footnotesize NLL\textsuperscript{$\downarrow$}}
& {\footnotesize AUROC\textsuperscript{$\uparrow$}}
& {\footnotesize ECE\textsuperscript{$\downarrow$}}
& {\footnotesize Brier\textsuperscript{$\downarrow$}}
& {\footnotesize NLL\textsuperscript{$\downarrow$}}
& {\footnotesize AUROC\textsuperscript{$\uparrow$}}
& {\footnotesize ECE\textsuperscript{$\downarrow$}}
& {\footnotesize Brier\textsuperscript{$\downarrow$}}
& {\footnotesize NLL\textsuperscript{$\downarrow$}} \\
\midrule
HTC
& .653 & .289 & .316 & .989
& .482 & .289 & .337 & 1.333
& .531 & .212 & .270 & .909
& .672 & .131 & .239 & .737 \\
MICE-LR
& .675 & .125 & .232 & .670
& .502 & .299 & .346 & .948
& .585 & .399 & .379 & 1.027
& .653 & .271 & .302 & .837 \\
MICE-RF
& .629 & .046 & .229 & .651
& .503 & .132 & .268 & .732
& .520 & .243 & .289 & .787
& .607 & .136 & .250 & .693 \\
CCPS
& .420 & .405 & .429 & 1.335
& .453 & .261 & .324 & 1.368
& .524 & .220 & .277 & 1.793
& .491 & .256 & .320 & 1.791 \\
DACA
& .668 & \textbf{.027} & .223 & .638
& .509 & \textbf{.072} & .247 & .686
& .556 & .218 & .249 & .692
& .623 & .165 & .258 & .710 \\
$P(\mathrm{Sufficient})$
& .661 & .331 & .337 & 1.131
& .530 & .373 & .392 & 1.088
& .653 & .478 & .432 & 1.204
& .663 & .485 & .462 & 1.417 \\
\midrule
Ours
& \textbf{.800} & .054 & \textbf{.181} & \textbf{.542}
& \textbf{.621} & .087 & \textbf{.240} & \textbf{.673}
& \textbf{.677} & \textbf{.204} & \textbf{.243} & \textbf{.686}
& \textbf{.748} & \textbf{.099} & \textbf{.209} & \textbf{.605} \\
% 0.67655 0.20393 0.245760.68735
% 0.74805 0.09940 0.20940 0.60531
\bottomrule
\end{tabular}%
}
\end{table*}

\begin{table}[t]
\centering
\caption{Cross-model generalization: Llama-3.2-3B-Instruct with PPO .}
\label{tab:cross_family}
\fontsize{10}{11}\selectfont
\setlength{\tabcolsep}{1.25pt}
\renewcommand{\arraystretch}{1.0}
\resizebox{\linewidth}{!}{
\begin{tabular}{@{}lcccc@{\hspace{5pt}}cccc@{\hspace{5pt}}cccc@{\hspace{5pt}}cccc@{}}
\toprule
& \multicolumn{4}{c}{TriviaQA}
& \multicolumn{4}{c}{HotpotQA}
& \multicolumn{4}{c}{MuSiQue}
& \multicolumn{4}{c}{PopQA} \\
\cmidrule(lr){2-5}
\cmidrule(lr){6-9}
\cmidrule(lr){10-13}
\cmidrule(lr){14-17}
Method
& {\footnotesize AUROC\textsuperscript{$\uparrow$}}
& {\footnotesize ECE\textsuperscript{$\downarrow$}}
& {\footnotesize Brier\textsuperscript{$\downarrow$}}
& {\footnotesize NLL\textsuperscript{$\downarrow$}}
& {\footnotesize AUROC\textsuperscript{$\uparrow$}}
& {\footnotesize ECE\textsuperscript{$\downarrow$}}
& {\footnotesize Brier\textsuperscript{$\downarrow$}}
& {\footnotesize NLL\textsuperscript{$\downarrow$}}
& {\footnotesize AUROC\textsuperscript{$\uparrow$}}
& {\footnotesize ECE\textsuperscript{$\downarrow$}}
& {\footnotesize Brier\textsuperscript{$\downarrow$}}
& {\footnotesize NLL\textsuperscript{$\downarrow$}}
& {\footnotesize AUROC\textsuperscript{$\uparrow$}}
& {\footnotesize ECE\textsuperscript{$\downarrow$}}
& {\footnotesize Brier\textsuperscript{$\downarrow$}}
& {\footnotesize NLL\textsuperscript{$\downarrow$}} \\
\midrule
HTC
& .683 & .114 & .233 & .669
& .631 & \textbf{.071} & .202 & .596
& .688 & .103 & .090 & .313
& .611 & .211 & .274 & .896 \\
MICE-LR
& .730 & .047 & .202 & .593
& .663 & .108 & .208 & .597
& .651 & .071 & .085 & .300
& .700 & .098 & .218 & .622 \\
MICE-RF
& .715 & \textbf{.029} & .202 & .587
& .681 & .097 & .197 & .569
& .661 & .069 & .080 & .286
& .713 & .103 & .212 & .601 \\
CCPS
& .707 & .105 & .210 & .607
& .686 & .126 & .201 & .580
& .652 & .103 & .087 & .308
& .691 & .083 & .213 & .619 \\
DACA
& .772 & .204 & .237 & .943
& .583 & .152 & .234 & .661
& .682 & .394 & .265 & .753
& .752 & .297 & .319 & 1.076 \\
P(Sufficient)
& .609 & .135 & .254 & .703
& .652 & .285 & .323 & .982
& .564 & .354 & .222 & .634
& .615 & \textbf{.064} & .234 & .659 \\
\midrule
Ours
& \textbf{.833} & .047 & \textbf{.167} & \textbf{.503}
& \textbf{.746} & .074 & \textbf{.179} & \textbf{.528}
& \textbf{.744} & \textbf{.045} & \textbf{.074} & \textbf{.260}
& \textbf{.830} & .093 & \textbf{.174} & \textbf{.516} \\
% 0.83328 0.04656 0.16740 0.50269
% 0.74579 0.07391 0.17932 0.52760
% 0.74327 0.04476 0.07401 0.26016
% 0.82978 0.09305 0.17351 0.51643
\bottomrule
\end{tabular}
}
\end{table}

We evaluate cross-dataset generalization in Table~\ref{tab:cross_dataset}. We train the estimator on Base-LLM trajectories from TriviaQA and evaluate it on Agentic RL trajectories from 2WikiMultiHopQA, Bamboogle, and NQ, with TriviaQA serving as an in-domain reference. 
Our method achieves the highest AUROC and the lowest Brier score and NLL on all four datasets.  
It also achieves the lowest ECE on Bamboogle and NQ. These results demonstrate strong cross-dataset generalization.

\begin{table}[t]
\centering
\caption{Cross-model generalization: Qwen2.5-7B-Instruct with PPO.}
\label{tab:cross_size}
\fontsize{10}{11}\selectfont
\setlength{\tabcolsep}{1.25pt}
\renewcommand{\arraystretch}{1.0}
\resizebox{\linewidth}{!}{%
\begin{tabular}{@{}lcccc@{\hspace{5pt}}cccc@{\hspace{5pt}}cccc@{\hspace{5pt}}cccc@{}}
\toprule
& \multicolumn{4}{c}{TriviaQA}
& \multicolumn{4}{c}{HotpotQA}
& \multicolumn{4}{c}{MuSiQue}
& \multicolumn{4}{c}{PopQA} \\
\cmidrule(lr){2-5}
\cmidrule(lr){6-9}
\cmidrule(lr){10-13}
\cmidrule(lr){14-17}
Method
& {\footnotesize AUROC\textsuperscript{$\uparrow$}}
& {\footnotesize ECE\textsuperscript{$\downarrow$}}
& {\footnotesize Brier\textsuperscript{$\downarrow$}}
& {\footnotesize NLL\textsuperscript{$\downarrow$}}
& {\footnotesize AUROC\textsuperscript{$\uparrow$}}
& {\footnotesize ECE\textsuperscript{$\downarrow$}}
& {\footnotesize Brier\textsuperscript{$\downarrow$}}
& {\footnotesize NLL\textsuperscript{$\downarrow$}}
& {\footnotesize AUROC\textsuperscript{$\uparrow$}}
& {\footnotesize ECE\textsuperscript{$\downarrow$}}
& {\footnotesize Brier\textsuperscript{$\downarrow$}}
& {\footnotesize NLL\textsuperscript{$\downarrow$}}
& {\footnotesize AUROC\textsuperscript{$\uparrow$}}
& {\footnotesize ECE\textsuperscript{$\downarrow$}}
& {\footnotesize Brier\textsuperscript{$\downarrow$}}
& {\footnotesize NLL\textsuperscript{$\downarrow$}} \\
\midrule
HTC
& .717 & .355 & .353 & 2.684
& .647 & .152 & .246 & .710
& .794 & .149 & .161 & .477
& .697 & .075 & .220 & .630 \\
MICE-LR
& .641 & .064 & .220 & .639
& .614 & .096 & .236 & .657
& .673 & .041 & .140 & .440
& .581 & .068 & .242 & .674 \\
MICE-RF
& .669 & \textbf{.036} & .209 & .607
& .629 & .087 & .228 & .638
& .645 & .061 & .146 & .456
& .652 & \textbf{.059} & .227 & .639 \\
CCPS
& .655 & .059 & .202 & .587
& .659 & \textbf{.078} & .224 & .631
& .672 & .087 & .147 & .458
& .625 & .146 & .253 & .697 \\
DACA
& .725 & .265 & .268 & 2.005
& .657 & .372 & .379 & 1.823
& .697 & .315 & .314 & 1.155
& .676 & .389 & .387 & 2.679 \\
P(Sufficient)
& .750 & .258 & .267 & 1.603
& .704 & .363 & .371 & 2.249
& .730 & .369 & .362 & 1.991
& .691 & .380 & .381 & 2.487 \\
\midrule
Ours
& \textbf{.822} & .112 & \textbf{.180} & \textbf{.539}
& \textbf{.769} & .081 & \textbf{.197} & \textbf{.569}
& \textbf{.805} & \textbf{.038} & \textbf{.121} & \textbf{.382}
& \textbf{.799} & .103 & \textbf{.193} & \textbf{.567} \\
\bottomrule
\end{tabular}%
}
\end{table}

\subsubsection{Generalization across Models}
We evaluate generalization across different model families using  Llama-3.2-3B-Instruct (Table~\ref{tab:cross_family}). 
We retrain the confidence estimator on the new model, while keeping all other configurations fixed. 
On Llama-3.2-3B-Instruct, we achieve the best AUROC, Brier score, and NLL on TriviaQA, HotpotQA, MuSiQue, and PopQA, with AUROC gains of 5.6--7.8 percentage points over the strongest baseline.
% We also experiment with different model scales using Qwen2.5-7B-Instruct in Appendix~\ref{app:model_generation}. 
% The results show that our method maintains strong performance even on larger models.
We also evaluate generalization across model scales using Qwen2.5-7B-Instruct (Table~\ref{tab:cross_size}) to test our metod. 
Under the 7B backbone, StaR-TTA achieves the highest AUROC on all four datasets, exceeding the strongest baseline by 1.1--10.2 percentage points, and obtains the lowest Brier score and NLL. 
These results demonstrate the effectiveness of our method across LLM scales and families.

\subsubsection{Generalization across RL Algorithms}

\begin{table}[h]
\centering
\caption{Generalization across RL algorithms: Qwen2.5-3B with GRPO.}
\label{tab:cross_rl}
\fontsize{10}{11}\selectfont
\setlength{\tabcolsep}{1.25pt}
\renewcommand{\arraystretch}{1.0}
\resizebox{\linewidth}{!}{%
\begin{tabular}{@{}lcccc@{\hspace{5pt}}cccc@{\hspace{5pt}}cccc@{\hspace{5pt}}cccc@{}}
\toprule
& \multicolumn{4}{c}{TriviaQA}
& \multicolumn{4}{c}{HotpotQA}
& \multicolumn{4}{c}{MuSiQue}
& \multicolumn{4}{c}{PopQA} \\
\cmidrule(lr){2-5}
\cmidrule(lr){6-9}
\cmidrule(lr){10-13}
\cmidrule(lr){14-17}
Method
& {\footnotesize AUROC\textsuperscript{$\uparrow$}}
& {\footnotesize ECE\textsuperscript{$\downarrow$}}
& {\footnotesize Brier\textsuperscript{$\downarrow$}}
& {\footnotesize NLL\textsuperscript{$\downarrow$}}
& {\footnotesize AUROC\textsuperscript{$\uparrow$}}
& {\footnotesize ECE\textsuperscript{$\downarrow$}}
& {\footnotesize Brier\textsuperscript{$\downarrow$}}
& {\footnotesize NLL\textsuperscript{$\downarrow$}}
& {\footnotesize AUROC\textsuperscript{$\uparrow$}}
& {\footnotesize ECE\textsuperscript{$\downarrow$}}
& {\footnotesize Brier\textsuperscript{$\downarrow$}}
& {\footnotesize NLL\textsuperscript{$\downarrow$}}
& {\footnotesize AUROC\textsuperscript{$\uparrow$}}
& {\footnotesize ECE\textsuperscript{$\downarrow$}}
& {\footnotesize Brier\textsuperscript{$\downarrow$}}
& {\footnotesize NLL\textsuperscript{$\downarrow$}} \\
\midrule
HTC
& .663 & .321 & .331 & 1.171
& .422 & .245 & .274 & 1.039
& .551 & .844 & .767 & 2.479
& .636 & .579 & .570 & 2.613 \\
MICE-LR
& .640 & .150 & .251 & .716
& .624 & .149 & .230 & .657
& .567 & .051 & .052 & .217
& .524 & .267 & .327 & .890 \\
MICE-RF
& .564 & .080 & .244 & .681
& .605 & \textbf{.068} & .204 & .593
& .587 & .079 & .057 & .234
& .521 & .112 & .254 & .701 \\
CCPS
& .425 & .536 & .546 & 4.192
& .437 & .259 & .275 & 1.661
& .483 & .210 & .104 & .434
& .467 & .318 & .354 & 2.919 \\
DACA
& .692 & .093 & .232 & .655
& .611 & .244 & .261 & .716
& .642 & .465 & .263 & .718
& .697 & .134 & .252 & .696 \\
P(Sufficient)
& .647 & .368 & .370 & 1.343
& .632 & .631 & .604 & 2.091
& .613 & .840 & .768 & 2.532
& .671 & .521 & .504 & 1.713 \\
\midrule
Ours
& \textbf{.756} & \textbf{.052} & \textbf{.199} & \textbf{.585}
& \textbf{.724} & .114 & \textbf{.192} & \textbf{.567}
& \textbf{.676} & \textbf{.042} & \textbf{.048} & \textbf{.199}
& \textbf{.750} & \textbf{.097} & \textbf{.204} & \textbf{.596} \\
% 0.75636 0.05146 0.19916 0.58533
% 0.72357 0.11417 0.19197 0.56653
% 0.67627 0.04173 0.04751 0.19906
% 0.75022 0.09653 0.20403 0.59602
\bottomrule
\end{tabular}%
}
\end{table}

Finally, we verify that our findings are not specific to the PPO algorithm used for Agentic RL training.
Table~\ref{tab:cross_rl} reports results on GRPO algorithms, while keeping the same Base LLM and evaluation protocol.
We achieve the best AUROC on all four datasets ($0.756$ on TriviaQA, $0.724$ on HotpotQA, $0.676$ on MuSiQue, and $0.750$ on PopQA), improving over the strongest baseline by $3.4$ to $9.2$ points, and additionally attain the best Brier and NLL on HotpotQA, MuSiQue, and PopQA.
These results confirm that trajectory shift is a general phenomenon of RL training and that trajectory-based confidence estimation consistently benefits across different RL algorithms.

\subsection{Ablation Studies}

\subsubsection{Effect of Test-Time Adaption}

\begin{wraptable}[9]{r}
{0.45\columnwidth}
\vspace{-5.9em}
\centering
\scriptsize
\setlength{\tabcolsep}{2pt}
\renewcommand{\arraystretch}{1.05}
\begin{tabular}{@{}lrrrrr@{}}
\toprule
    Dataset & Setting & AUROC$\uparrow$ & ECE$\downarrow$ & Brier$\downarrow$ & NLL$\downarrow$ \\
    \midrule
    TriviaQA & w/o TTA & 0.8018 & 0.0455 & 0.1794 & 0.5386 \\
             & w/ TTA  & 0.8000 & 0.0535 & 0.1808 & 0.5421 \\
    \midrule
    HotpotQA & w/o TTA & 0.7112 & 0.0349 & 0.1997 & 0.5828 \\
             & w/ TTA  & 0.7154 & 0.0409 & 0.1996 & 0.5824 \\
    \midrule
    MuSiQue  & w/o TTA & 0.6296 & 0.1212 & 0.1478 & 0.5503 \\
             & w/ TTA  & 0.6721 & 0.1198 & 0.1450 & 0.5316 \\
    \midrule
    PopQA    & w/o TTA & 0.6714 & 0.0600 & 0.2273 & 0.6464 \\
             & w/ TTA  & 0.6956 & 0.0610 & 0.2226 & 0.6354 \\
    \midrule
    Average  & w/o TTA & 0.7035 & 0.0654 & 0.1885 & 0.5795 \\
             & w/ TTA  & 0.7208 & 0.0688 & 0.1870 & 0.5729 \\
    \bottomrule
\end{tabular}
\caption{Effect of removing TTA. }
\label{tab:tta_removal}
\end{wraptable}

We compare the full method with a variant that removes TTA and directly maps the original student scores through the same frozen calibrator. 
Table~\ref{tab:tta_removal} reports the actual metrics for both settings. 
TTA improves AUROC on HotpotQA, MuSiQue, and PopQA, while slightly reducing it on TriviaQA. 
The average AUROC increases by approximately 1.73 percentage points.
Average Brier score and NLL also improve, whereas average ECE increases slightly. 
These results show that TTA mainly benefits confidence ranking, with mixed effects on calibration.
We also report ablation results on external descriptors $d_i^m$, stability vectors $c_i^m$, $q^{\mathrm{tta}}$, and our training loss in Appendix~\ref{app:trajectory_ablations}, where we find removing any of them will decrease the performance.

\subsubsection{TTA Sample Efficiency}

\begin{wraptable}[13]{r}{0.43\columnwidth}
\vspace{-1em}
\centering
\scriptsize
\setlength{\tabcolsep}{1.8pt}
\renewcommand{\arraystretch}{1.08}
\begin{tabular}{@{}rccccc@{}}
\toprule
$N$ & Average& Trivia & Hotpot & MuSiQue & PopQA \\
\midrule
0 & 0.7035 & 0.8018 & 0.7112 & 0.6296 & 0.6714 \\
1 & 0.7035 & 0.8018 & 0.7112 & 0.6296 & 0.6714 \\
2 & 0.6306 & 0.7537 & 0.5125 & 0.6383 & 0.6181 \\
4 & 0.6680 & 0.7828 & 0.5744 & 0.6671 & 0.6477 \\
8 & 0.7089 & 0.7911 & 0.6839 & 0.6683 & 0.6924 \\
12 & 0.7072 & 0.7914 & 0.6893 & 0.6703 & 0.6779 \\
16 & 0.7056 & 0.7943 & 0.6811 & 0.6734 & 0.6738 \\
32 & 0.7120 & 0.8002 & 0.7112 & 0.6709 & 0.6658 \\
64 & 0.7155 & 0.8014 & 0.7151 & 0.6734 & 0.6723 \\
All & 0.7208 & 0.8000 & 0.7154 & 0.6721 & 0.6956 \\
\bottomrule
\end{tabular}
\caption{AUROC under different TTA sample sizes.}
\label{tab:tta_samples}
\vspace{-0.8em}
\end{wraptable}

% Our TTA procedure uses stability-based pairwise rankings to adjust confidence scores. More unlabeled target questions can provide more informative comparisons and make these corrections more reliable. We therefore vary the number of questions used for adaptation, as shown in Table~\ref{tab:tta_samples}. A single question provides no pairwise comparisons and yields almost no improvement, while gains emerge with 4--8 questions. Average AUROC increases from 0.7035 without TTA to 0.7056 with 16 questions and 0.7208 using all available questions. 
% These results suggest that additional unlabeled examples improve the reliability of ranking-based adaptation, with diminishing gains as the sample budget increases.

Our TTA procedure uses stability-based pairwise rankings to adjust confidence scores. 
More unlabeled target questions can provide more informative comparisons and make results more reliable. 
We therefore vary the number of questions used for adaptation, as shown in Table~\ref{tab:tta_samples}. 
With only a small number of adaptation samples (e.g., $N=2$ or $N=4$), the results become unstable and even decrease. 
Performance gradually recovers and stabilizes as we increase the number of unlabeled questions for adaptation. 
The average AUROC rises from 0.7035 without TTA to 0.7155 with 64 questions.
We can find that only a small number of samples (e.g., 64) during TTA is sufficient.
These results suggest that additional unlabeled examples improve the reliability of ranking-based adaptation, with diminishing gains as the sample budget increases.
We further analyze the additional cost introduced by our method in Appendix~\ref{app:cost_analysis}.
The results show that the stability signal only introduces an extra token overhead of merely 0.8\%.

\section{Conclusion}

We studied why confidence estimators trained on Base LLM trajectories degrade after Agentic RL. 
We find that RL-induced trajectory shifts account for a substantial fraction of the degradation.
Despite this shift, trajectory stability remains predictive of answer correctness in both Base and RL models. 
This preserved relationship motivates StaR-TTA, a label-free method that adjusts confidence scores on Agentic RL trajectories at test time. 
Across four datasets, StaR-TTA achieves an average AUROC of \(0.721\). 
It also generalizes across datasets, model families, and PPO- or GRPO-trained agents. 
These results suggest that StaR-TTA may provide a useful starting point for further research on confidence estimation in Agentic-RL trained LLM agents.

\subsection*{AI use statement}

In this work, we used generative AI tools for providing feedback on research  methodology.
Additionally, we used generative AI tools for edit a research paper to improve readability. 
We have reviewed all AI-assisted work. 
We checked LLM-generated feedback and we manually polished the content generated by the LLM. 
We take responsibility for the final content of this work, including text, claims or artifacts produced with the aid of generative AI.

\bibliography{iclr2027_conference}
\bibliographystyle{iclr2027_conference}

\appendix

\section{Appendix}

\begin{table}[htbp]
  \centering
  \small
  \setlength{\tabcolsep}{5pt}
  \caption{Independent ablations of Scorer input features. Each ablation removes only one feature group.}
  \label{tab:ablation_student_inputs}
  \begin{tabular}{llcccc}
    \toprule
    Dataset & Setting & AUROC$\uparrow$ & ECE$\downarrow$ & Brier$\downarrow$ & NLL$\downarrow$ \\
    \midrule
    TriviaQA & w/o Stability input $c^m$ & 0.7413 & 0.1529 & 0.2250 & 0.6408 \\
             & w/o Statistics $d^m$      & 0.7871 & 0.0473 & 0.1850 & 0.5526 \\
             & Ours                & 0.8000 & 0.0535 & 0.1808 & 0.5421 \\
    \midrule
    HotpotQA & w/o Stability input $c^m$ & 0.6170 & 0.1222 & 0.2337 & 0.6679 \\
             & w/o Statistics $d^m$      & 0.6992 & 0.0724 & 0.2079 & 0.6030 \\
             & Ours                & 0.7154 & 0.0409 & 0.1996 & 0.5824 \\
    \midrule
    MuSiQue  & w/o Stability input $c^m$ & 0.6578 & 0.1241 & 0.1478 & 0.5422 \\
             & w/o Statistics $d^m$      & 0.6659 & 0.1061 & 0.1411 & 0.5024 \\
             & Ours                & 0.6721 & 0.1198 & 0.1450 & 0.5316 \\
    \midrule
    PopQA    & w/o Stability input $c^m$ & 0.6645 & 0.2693 & 0.3009 & 0.8716 \\
             & w/o Statistics $d^m$     & 0.6675 & 0.0311 & 0.2259 & 0.6438 \\
             & Ours                & 0.6956 & 0.0610 & 0.2226 & 0.6354 \\
    \bottomrule
  \end{tabular}
\end{table}

\subsection{Ablation Studies}
\label{app:trajectory_ablations}

To analyze the contribution of different components in our framework, we conduct ablation studies on both the trajectory representation and the test-time adaptation strategy. 
For trajectory feature construction, we incorporate three types of signals: internal representations $h^m$, external descriptors $d^m$, and stability vectors $c^m$. 
Since using hidden-layer representations $h^m$ as trajectory features is a natural choice, we further investigate whether the additional stability and statistical signals provide complementary benefits.

During TTA, we construct $q^{\mathrm{tta}}$ to guide the ranking-based adaptation, which leverages both the stability vector $c^m$ and the agreement between Agentic-RL and Base trajectories. 
This differs from the training stage, where only $c^m$ is used. 
Therefore, we examine whether introducing the Base--RL agreement brings additional improvements.

During training, we use correctness and ranking losses based on ground-truth labels, along with two corresponding losses constructed from label-free stability signals. 
We further conduct ablation experiments to assess the individual contributions of these four loss terms.

% Requires \usepackage{booktabs}

\subsubsection{Scorer Input Features}

We independently remove stability features, and the 24-dimensional interaction statistics from the TCS input. 
Each ablation retains the other input features and follows the same source-training and calibration protocol. 
Stability signals remain available to the SRM, training losses, and TTA. 
Table~\ref{tab:ablation_student_inputs} shows that removing stability inputs worsens all four metrics on every dataset. 
Removing interaction statistics also lowers AUROC on all four datasets, although ECE improves on three datasets and MuSiQue obtains lower Brier score and NLL. 
These results support retaining both feature groups, while showing that their benefits vary across metrics.

\subsubsection{Base--RL Agreement during Adaptation}

\begin{table}[htbp]
  \centering
  \small
  \setlength{\tabcolsep}{5pt}
  \caption{Effect of removing Base--RL JS agreement from TTA.}
  \label{tab:ablation_js_agreement}
  \begin{tabular}{llcccc}
    \toprule
    Dataset & Setting & AUROC$\uparrow$ & ECE$\downarrow$ & Brier$\downarrow$ & NLL$\downarrow$ \\
    \midrule
    TriviaQA & w/o JS agreement & 0.8000 & 0.0521 & 0.1816 & 0.5441 \\
             & Ours             & 0.8000 & 0.0535 & 0.1808 & 0.5421 \\
    \midrule
    HotpotQA & w/o JS agreement & 0.7067 & 0.0402 & 0.2022 & 0.5885 \\
             & Ours             & 0.7154 & 0.0409 & 0.1996 & 0.5824 \\
    \midrule
    MuSiQue  & w/o JS agreement & 0.6743 & 0.1198 & 0.1449 & 0.5310 \\
             & Ours             & 0.6721 & 0.1198 & 0.1450 & 0.5316 \\
    \midrule
    PopQA    & w/o JS agreement & 0.6927 & 0.0633 & 0.2238 & 0.6383 \\
             & Ours             & 0.6956 & 0.0610 & 0.2226 & 0.6354 \\
    \bottomrule
  \end{tabular}
\end{table}

We remove Base--RL JS agreement from the target quality signal and use only the mean RL stability, $q^{\mathrm{tta}}=S_{\mathrm{raw3}}$, where  $\mathcal S_{\mathrm{raw3}}$ is the equal-weight average of the stability vector.
The SRM, confidence calibrator, and other TTA settings remain unchanged. 
As shown in Table~\ref{tab:ablation_js_agreement}, removing agreement lowers AUROC on HotpotQA and PopQA, while leaving TriviaQA nearly unchanged and slightly improving MuSiQue. 
Brier score and NLL increase on TriviaQA, HotpotQA, and PopQA. 
These results indicate that Base--RL agreement provides useful information beyond RL stability, although its benefit varies across datasets.

\subsubsection{Training Loss Ablation}
\label{app:ablation_loss}
\begin{table}[t]
\centering
\caption{Ablation of individual training losses on PopQA. Average degradation denotes the mean relative degradation across the four metrics compared with the full method.}
\label{tab:loss_ablation}
\small
\setlength{\tabcolsep}{5pt}
\renewcommand{\arraystretch}{1.0}
\begin{tabular}{@{}lccccc@{}}
\toprule
Setting
& AUROC$\uparrow$
& ECE$\downarrow$
& Brier$\downarrow$
& NLL$\downarrow$
& \shortstack{Average degradation\\(\%)$\downarrow$} \\
\midrule
w/o Label BCE
& 0.6952 & 0.1151 & 0.2372 & 0.6659 & 
25.03 \\
w/o Label Ranking
& 0.6886 & 0.0647 & 0.2248 & 0.6408 & 2.23 \\
w/o Stability Ranking
& 0.7180 & 0.1131 & 0.2250 & 0.6457 & 21.22 \\
w/o SRM BCE
& 0.6972 & 0.0634 & 0.2223 & 0.6345 & 0.86 \\
\midrule
Full method
& 0.6956 & 0.0610 & 0.2226 & 0.6354 & 0.00 \\
\bottomrule
\end{tabular}
\end{table}

To examine the contribution of each training loss, we remove one loss term at a time and evaluate the resulting variants on PopQA.
Table~\ref{tab:loss_ablation} reports the average relative degradation across the four metrics, using the full method as the reference.
Removing Label BCE produces the largest average degradation of $25.03\%$, followed by removing Stability Ranking at $21.22\%$.
These results suggest that combining direct correctness supervision with stability-based ranking supervision is important for the aggregate performance of the confidence estimator.
Removing Label Ranking and SRM BCE leads to smaller average degradations of $2.23\%$ and $0.86\%$, respectively, indicating more modest contributions under this aggregate measure.
Although the effects differ in magnitude, all ablated variants exhibit positive average degradation, supporting the combined training objective in this setting.

\subsection{Direct Confidence Estimation from Stability}
\label{app: simple_baseline}

We investigate whether stability signals alone suffice for confidence estimation, potentially providing a simpler alternative to our full framework.
We compare five stability-only predictors, each using the three-dimensional stability vector with no hidden layers: logistic regression, probit regression, complementary log-log regression, a sigmoid output trained with Brier loss, and ridge regression with predictions clipped to the probability range. 
Each predictor is trained on labeled Base LLM trajectories, with regularization selected by Base-validation NLL, and evaluated on the RL test trajectories.
Table~\ref{tab:stability_only_average} reports equal-weight averages over TriviaQA, HotpotQA, MuSiQue, and PopQA. Our method achieves the best average performance on all four metrics, supporting the integration of stability signals with trajectory information.

\begin{table}[htbp]
  \centering
  \small
  \setlength{\tabcolsep}{4pt}
  \caption{Confidence estimation averaged equally across four datasets. }
  \label{tab:stability_only_average}
  \begin{tabular}{lcccc}
    \toprule
    Method & AUROC$\uparrow$ & ECE$\downarrow$
    & Brier$\downarrow$ & NLL$\downarrow$ \\
    \midrule
    Logistic regression
    & 0.6803 & 0.0783 & 0.1950 & 0.5811 \\
    Probit regression
    & 0.6807 & 0.0781 & 0.1958 & 0.5855 \\
    Complementary log-log
    & 0.6802 & 0.0791 & 0.1953 & 0.5825 \\
    Sigmoid (Brier loss)
    & 0.6821 & 0.0736 & 0.1942 & 0.5775 \\
    Clipped ridge regression
    & 0.6803 & 0.0699 & 0.1944 & 0.5782 \\
    \midrule
    Ours
    & \textbf{0.7208} & \textbf{0.0688}
    & \textbf{0.1870} & \textbf{0.5729} \\
    \bottomrule
  \end{tabular}
\end{table}

\begin{figure}[htbp]
  \centering
  \includegraphics[width=\textwidth]{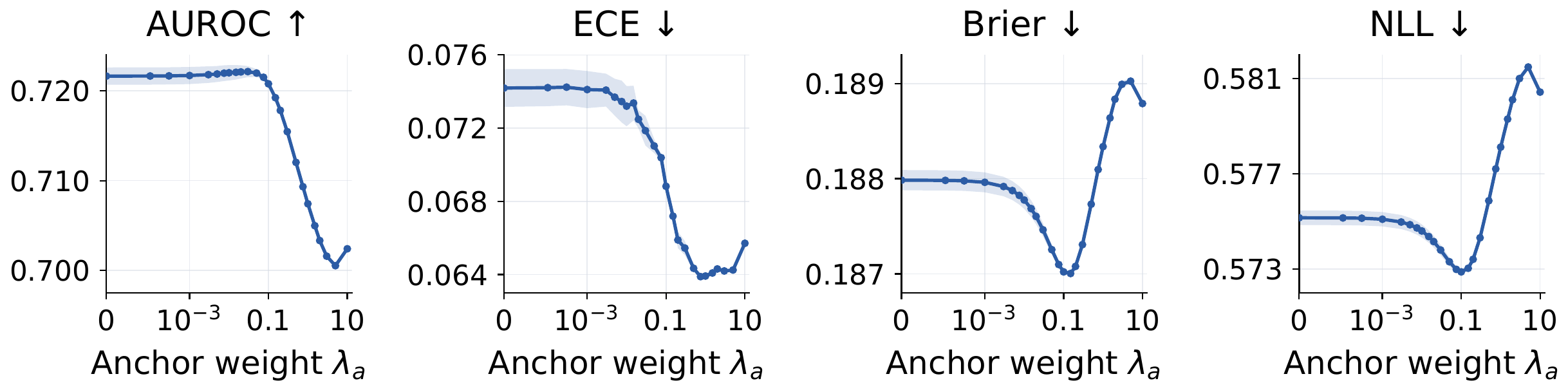}
  \caption{Sensitivity to the anchoring weight $\lambda_a$.
  Curves show the equal-weight average over four datasets. }
  \label{fig:anchor_weight}
\end{figure}

\subsection{Hyperparameter Analysis}

\subsubsection{Robustness to the Anchoring Weight}
\label{app:anchor_weight}

During TTA, the corrector is trained with a ranking loss and an anchoring loss. 
The ranking loss encourages scores to follow the stability-based ordering, while the anchoring loss limits their deviation from the initial student scores. 
Their balance is controlled by $\lambda_a$ in $\mathcal{L}_{\mathrm{tta}}=\mathcal{L}_{\mathrm{rank}}+\lambda_a\mathcal{L}_{\mathrm{anchor}}$. 
A smaller weight allows stronger corrections, whereas a larger weight favors the original predictions. 
Figure~\ref{fig:anchor_weight} varies $\lambda_a$ while keeping the other settings fixed on training set. 
AUROC peaks near $0.03$, while Brier score and NLL reach their lowest values around $0.1$--$0.15$.
Larger weights generally reduce AUROC and eventually worsen Brier score and NLL, although ECE remains lower. 
We use $\lambda_a=0.1$ as a compromise between ranking and probability quality.

\subsubsection{Robustness to Training Loss Weights}
\label{app:train_hy}
In our experiments, we set $\lambda_{\mathrm{h}}=0.5$, $\lambda_r=1.0$, and $\lambda_s=0.5$.
To assess robustness to these hyperparameters, we substantially vary each weight individually: doubling $\lambda_{\mathrm{h}}$ to $1.0$, reducing $\lambda_r$ to $0.2$, and doubling $\lambda_s$ to $1.0$, while keeping all other settings fixed.
As shown in Table~\ref{tab:loss_weight_robustness}, the mean absolute change across the three perturbations and four evaluation metrics is below $0.003$, suggesting that our method is relatively insensitive to the tested variations in loss weights.

\begin{table}[htbp]
  \centering
  \small
  \setlength{\tabcolsep}{4pt}
  \caption{Robustness to training loss weightss.}
  \label{tab:loss_weight_robustness}
  \begin{tabular}{cccccccc}
    \toprule
    $\lambda_{\mathrm{h}}$ & $\lambda_r$ & $\lambda_s$
    & AUROC$\uparrow$ & ECE$\downarrow$
    & Brier$\downarrow$ & NLL$\downarrow$ &  Change \\
    \midrule
    0.5 & 1.0 & 0.5 & 0.7208 & 0.0688 & 0.1870 & 0.5729 & - \\
    \midrule
    1.0$\uparrow$ & 1.0 & 0.5 & 0.7227 & 0.0656 & 0.1860 & 0.5662 & $-0.0023$ \\
    0.5 & 0.2$\downarrow$ & 0.5 & 0.7229 & 0.0711 & 0.1870 & 0.5785 & $+0.0025$ \\
    0.5 & 1.0 & 1.0$\uparrow$ & 0.7193 & 0.0710 & 0.1879 & 0.5786 & $+0.0018$ \\
    \bottomrule
  \end{tabular}
\end{table}

\begin{table}[t]
  \centering
  \footnotesize
  \setlength{\tabcolsep}{3pt}
  \caption{Confidence estimation on Base and RL test trajectories.}
  \label{tab:base_rl_test}
  \begin{tabular}{lcccccccc}
    \toprule
    & \multicolumn{4}{c}{Base trajectories}
    & \multicolumn{4}{c}{RL trajectories} \\
    \cmidrule(lr){2-5}\cmidrule(lr){6-9}
    Method
    & AUROC$\uparrow$ & ECE$\downarrow$ & Brier$\downarrow$ & NLL$\downarrow$
    & AUROC$\uparrow$ & ECE$\downarrow$ & Brier$\downarrow$ & NLL$\downarrow$ \\
    \midrule
    \multicolumn{9}{l}{\textbf{TriviaQA} } \\
    \midrule
    DACA$^\dagger$
    & 0.6979 & 0.2257 & 0.2625 & 0.7223
    & 0.6675 & 0.0265 & 0.2232 & 0.6376 \\
    P(Sufficient)$^\dagger$
    & 0.6815 & 0.4786 & 0.4497 & 1.3340
    & 0.6605 & 0.3306 & 0.3372 & 1.1313 \\
    HTC
    & 0.7976 & 0.0162 & 0.1765 & 0.5391
    & 0.6529 & 0.2887 & 0.3161 & 0.9893 \\
    MICE-LR
    & 0.7723 & 0.0223 & 0.1871 & 0.5540
    & 0.6754 & 0.1249 & 0.2324 & 0.6703 \\
    MICE-RF
    & 0.7885 & 0.0301 & 0.1812 & 0.5312
    & 0.6290 & 0.0461 & 0.2294 & 0.6506 \\
    CCPS
    & 0.7868 & 0.0224 & 0.1813 & 0.5370
    & 0.4197 & 0.4045 & 0.4286 & 1.3352 \\
    Ours
    & 0.7840 & 0.0210 & 0.1806 & 0.5371
    & 0.8000 & 0.0535 & 0.1808 & 0.5421 \\
    \midrule
    \multicolumn{9}{l}{\textbf{HotpotQA}} \\
    \midrule
    DACA$^\dagger$
    & 0.6517 & 0.3868 & 0.2737 & 0.7441
    & 0.6128 & 0.1547 & 0.2432 & 0.6793 \\
    P(Sufficient)$^\dagger$
    & 0.6496 & 0.6698 & 0.5740 & 1.6223
    & 0.6363 & 0.5262 & 0.4983 & 1.5509 \\
    HTC
    & 0.7397 & 0.0304 & 0.1104 & 0.3619
    & 0.5067 & 0.1750 & 0.2684 & 0.7382 \\
    MICE-LR
    & 0.7268 & 0.0321 & 0.1146 & 0.3726
    & 0.6403 & 0.1122 & 0.2341 & 0.6623 \\
    MICE-RF
    & 0.7530 & 0.0504 & 0.1101 & 0.3571
    & 0.6149 & 0.0475 & 0.2190 & 0.6292 \\
    CCPS
    & 0.7329 & 0.0347 & 0.1098 & 0.3634
    & 0.4381 & 0.2048 & 0.2842 & 0.9195 \\
    Ours
    & 0.7482 & 0.0293 & 0.1084 & 0.3551
    & 0.7154 & 0.0409 & 0.1996 & 0.5824 \\
    \midrule
    \multicolumn{9}{l}{\textbf{MuSiQue}} \\
    \midrule
    DACA$^\dagger$
    & 0.6813 & 0.4619 & 0.2616 & 0.7165
    & 0.6253 & 0.3103 & 0.2285 & 0.6497 \\
    P(Sufficient)$^\dagger$
    & 0.6402 & 0.7456 & 0.6080 & 1.6726
    & 0.6475 & 0.6425 & 0.5505 & 1.5842 \\
    HTC
    & 0.7055 & 0.0186 & 0.0314 & 0.1398
    & 0.7209 & 0.0808 & 0.1317 & 0.4229 \\
    MICE-LR
    & 0.7026 & 0.0174 & 0.0331 & 0.1428
    & 0.6231 & 0.0540 & 0.1350 & 0.4434 \\
    MICE-RF
    & 0.6618 & 0.0409 & 0.0368 & 0.1599
    & 0.6101 & 0.0637 & 0.1360 & 0.4420 \\
    CCPS
    & 0.6640 & 0.0337 & 0.0345 & 0.1540
    & 0.4936 & 0.1140 & 0.1485 & 0.5337 \\
    Ours
    & 0.7610 & 0.0088 & 0.0303 & 0.1293
    & 0.6721 & 0.1198 & 0.1450 & 0.5316 \\
    \midrule
    \multicolumn{9}{l}{\textbf{PopQA}} \\
    \midrule
    DACA$^\dagger$
    & 0.6420 & 0.2220 & 0.2519 & 0.6987
    & 0.6804 & 0.1098 & 0.2438 & 0.6806 \\
    P(Sufficient)$^\dagger$
    & 0.6615 & 0.5181 & 0.4706 & 1.3192
    & 0.6405 & 0.4589 & 0.4459 & 1.3821 \\
    HTC
    & 0.7439 & 0.0658 & 0.1766 & 0.5338
    & 0.6779 & 0.3277 & 0.3337 & 0.9366 \\
    MICE-LR
    & 0.7361 & 0.0477 & 0.1830 & 0.5431
    & 0.5953 & 0.2816 & 0.3211 & 0.8815 \\
    MICE-RF
    & 0.7937 & 0.0384 & 0.1636 & 0.4918
    & 0.5903 & 0.1186 & 0.2539 & 0.7015 \\
    CCPS
    & 0.7432 & 0.0392 & 0.1803 & 0.5307
    & 0.4249 & 0.3060 & 0.3573 & 1.4068 \\
    Ours
    & 0.7311 & 0.0643 & 0.1791 & 0.5347
    & 0.6956 & 0.0610 & 0.2226 & 0.6354 \\
    \bottomrule
  \end{tabular}
  \par\smallskip

\end{table}

\subsection{Confidence Estimation on Base and RL Trajectories}
\label{app:base_rl_test_comparison}

Table~\ref{tab:base_rl_test} compares confidence estimation on held-out Base and RL trajectories. We evaluate our source-trained TCS and frozen calibrator on Base trajectories, and report the full TTA method on RL trajectories.
Several baselines achieve strong performance on Base trajectories but deteriorate substantially when transferred to RL trajectories. 
For example, CCPS achieves AUROC values of $0.7868$ on TriviaQA and $0.7329$ on HotpotQA before transfer, which drop to $0.4197$ and $0.4381$ after transfer. 
Similarly, MICE-LR and MICE-RF exhibit lower AUROC on RL trajectories across all four datasets, despite their strong performance on Base trajectories.
These comparisons show existing estimators can successfully estimate confidence on held-out trajectories from the Base model, yet their learned confidence signals become less reliable after Agentic RL changes the trajectory distribution.

\subsection{Cost Analysis}
\label{app:cost_analysis}

We distinguish standard source training and confidence scoring from the additional costs of stability estimation and test-time adaptation.
At test time, we freeze TCS and CC then update only a linear corrector using unlabeled target examples.
Paired Base and RL trajectories are required only for training the corrector.
Once corrector is trained, confidence scoring uses RL trajectories alone, without running the Base model on subsequent questions.
Table~\ref{tab:tta_samples} shows that 64 adaptation examples improve average AUROC from 0.7035 to 0.7155, compared with 0.7208 using all available examples.
Thus, a small adaptation subset can provide useful gains while limiting the cost of Base-model execution.
Other confidence estimators may need to perform retraining on a large collection of RL trajectories (e.g., tens of thousands).
Without this retraining, they undergo substantial performance degradation.

For stability estimation, we fix each trajectory prefix and sample five short final answers, avoiding repeated reasoning and tool calls.
We estimate the additional answer tokens cost using 100 randomly selected Base trajectories per dataset.
As reported in Table~\ref{tab:cost_stability}, the five answers contain 46.1 tokens on average, or approximately 9.2 tokens per answer, compared with 5,824 tokens in the original trajectory.
The resulting 0.8\% ratio measures the token increment from answer resampling.

% \subsection{Cost Analysis}
% \label{app:cost_analysis}

% Our method has the same training requirement as other confidence estimators on base LLM trajectories.
% After the Base LLM is adapted with Agentic RL, we performs label-free test-time adaptation without retraining our estimator.
% Specifically, we update a corrector, which is merely a linear layer, using a very small number of samples to correct TCS.
% As shown in Table~\ref{tab:tta_samples}, only 32 or 64 unlabeled samples are sufficient.
% Other confidence estimators may need to be retrained on a large collection of RL trajectories (e.g., tens of thousands). 
% Without this retraining, they undergo substantial performance degradation.

% The additional cost of our method comes from computing the stability signal.
% For each trajectory, we keep the original trajectories fixed and only resample the answer for 5 times. 
% Therefore, the additional tokens cost introduced by our method is the total tokens of the five answers.
% We estimate this cost using 100 random selected Base trajectories from each dataset. 
% As reported in Table~\ref{tab:cost_stability}, the 5 sampled answers contain 46.1 tokens on average, corresponding to approximately 9.2 tokens per answer.
% The original trajectories require approximately 5,824  tokens on average.
% Thus, the additional answer decoding accounts for only 0.8\% of the full trajectory processing cost. 
% Therefore, the stability estimation introduces only a small overhead while avoiding regeneration of the trajectories.

\begin{table}[htbp]
  \centering
  \small
  \setlength{\tabcolsep}{8pt}
  \caption{Token usage of original trajectories and five sampled answers.
  }
  \label{tab:cost_stability}
  \begin{tabular}{lrrr}
    \toprule
    Dataset
    & \shortstack{Total original\\tokens}
    & \shortstack{Five answers\\tokens}
    & \shortstack{Five answers /\\Total original (\%)} \\
    \midrule
    TriviaQA & 5,034.5 & 45.0 & 0.9 \\
    HotpotQA & 5,289.3 & 45.8 & 0.9 \\
    MuSiQue  & 6,330.2 & 47.0 & 0.7 \\
    PopQA    & 6,642.5 & 46.8 & 0.7 \\
    \midrule
    Overall  & 5,824.1 & 46.1 & 0.8 \\
    \bottomrule
  \end{tabular}
\end{table}

\subsection{Experimental Details}
\label{app:experimental_protocol}

The source-training/source-validation/RL-test counts are 1,502/475/11,122 for TriviaQA, 1,486/478/6,687 for HotpotQA, 1,458/457/1,755 for MuSiQue, and 1,475/463/11,712 for PopQA.
As for 64-dimensional hidden representation, we use layers 28 and 36 for Qwen2.5-3B, and layers 24 and 28 for Qwen2.5-7B-Instruct and Llama-3.2-3B-Instruct.
The SRM uses three-input logistic regression.
The TCS is a two-branch MLP with two hidden layers per branch, whose scalar outputs are summed. 
It is trained using AdamW with learning rate $10^{-3}$, weight decay $10^{-4}$, batch size 128, and source-ranking threshold $0.3$.
The CC uses Platt scaling, $p=\sigma(as+b)$,
fitted on source OOF scores.
We sample five terminal answers with temperature $1.0$, top-$p=0.9$, top-$k=50$.
The corrector is a zero-initialized $288\to1$ linear layer which is trained for 20 Adam steps with learning rate $0.03$.
We judge whether the answer is correct by using exact match against any reference after lowercasing, removing punctuation and articles, and normalizing whitespace.

\end{document}